%% file: main_arxiv.tex
\documentclass[10pt,twocolumn,letterpaper]{article}

\usepackage[pagenumbers]{iccv} 


\definecolor{iccvblue}{rgb}{0.21,0.49,0.74}
\usepackage[pagebackref,breaklinks,colorlinks,allcolors=iccvblue]{hyperref}

\usepackage{booktabs}
\usepackage{arydshln}
\usepackage{colortbl}
\usepackage{xcolor}
\usepackage{multirow}
\usepackage{algorithm}
\usepackage{algorithmic}
\usepackage{makecell}
\usepackage{mdframed}
\usepackage{tcolorbox}
\usepackage{longtable}

\definecolor{task}{RGB}{139, 0, 0}
\definecolor{reward}{RGB}{46,139,87}
\definecolor{explore}{RGB}{218,165,32}
\definecolor{update}{RGB}{0, 0, 0}
\newcommand{\Model}{VCA\xspace}
\def\paperID{6771} 
\def\confName{ICCV}
\def\confYear{2025}
\makeatletter
\def\@fnsymbol#1{\ensuremath{\ifcase#1\or \dagger\or \ddagger\or
\mathsection\or \mathparagraph\or \|\or **\or \dagger\dagger
\or \ddagger\ddagger \else\@ctrerr\fi}}
\newcommand{\printfnsymbol}[1]{%
  \textsuperscript{\@fnsymbol{#1}}%
}
\makeatother
\title{VCA: Video Curious Agent for Long Video Understanding}

\author{Zeyuan Yang\thanks{ Equal Contribution}\\
University of Massachusetts, Amherst\\
{\tt zeyuanyang@umass.edu}
\and
Delin Chen\printfnsymbol{1}\\
University of Massachusetts, Amherst\\
{\tt delinchen@umass.edu}
\and
Xueyang Yu\\
University of Massachusetts, Amherst\\
{\tt xueyangyu@umass.edu}
\and
Maohao Shen\\
Massachusetts Institute of Technology\\
{\tt maohao@mit.edu}
\and
Chuang Gan\\
University of Massachusetts, Amherst\\
{\tt ganchuang@csail.mit.edu}
}

\begin{document}
\maketitle
\input{sec/0_abstract}    
\input{sec/1_intro}
\input{sec/2_related}
\input{sec/3_approach}
\input{sec/4_experiments}
\input{sec/5_analysis}

\input{sec/6_conclusion}
{
    \small
    \bibliographystyle{ieeenat_fullname}
    \bibliography{main}
}
\input{sec/X_suppl}

\end{document}

%% file: sec/0_abstract.tex
\begin{abstract}
Long video understanding poses unique challenges due to their temporal complexity and low information density. Recent works address this task by sampling numerous frames or incorporating auxiliary tools using LLMs, both of which result in high computational costs. In this work, we introduce a curiosity-driven video agent with self-exploration capability, dubbed as ``\Model''.
Built upon VLMs, \Model autonomously navigates video segments and efficiently builds a comprehensive understanding of complex video sequences.
Instead of directly sampling frames, \Model employs a tree-search structure to explore video segments and collect frames.
Rather than relying on external feedback or reward, \Model leverages VLM's self-generated intrinsic reward to guide its exploration, enabling it to capture the most crucial information for reasoning. Experimental results on multiple long video benchmarks demonstrate our approach’s superior effectiveness and efficiency.
\end{abstract}

%% file: sec/1_intro.tex
\section{Introduction}
\label{sec:intro}

There is a growing demand for developing long-form video understanding techniques~\cite{wu2021towards, soldan2022mad} capable of extracting valuable information from extended visual narratives, such as surveillance footage, documentaries, and instructional videos. Analyzing these videos~\cite{tapaswi2016movieqa, wu2024longvideobench}, which can range from minutes to hours, requires models that can process multimodal data and perform reasoning over extremely long sequences~\cite{li2024mvbench, bai2023qwen, wang2024qwen2}.

\begin{figure}[!ht]
    \centering
    \centerline{
    \includegraphics[width=\linewidth]{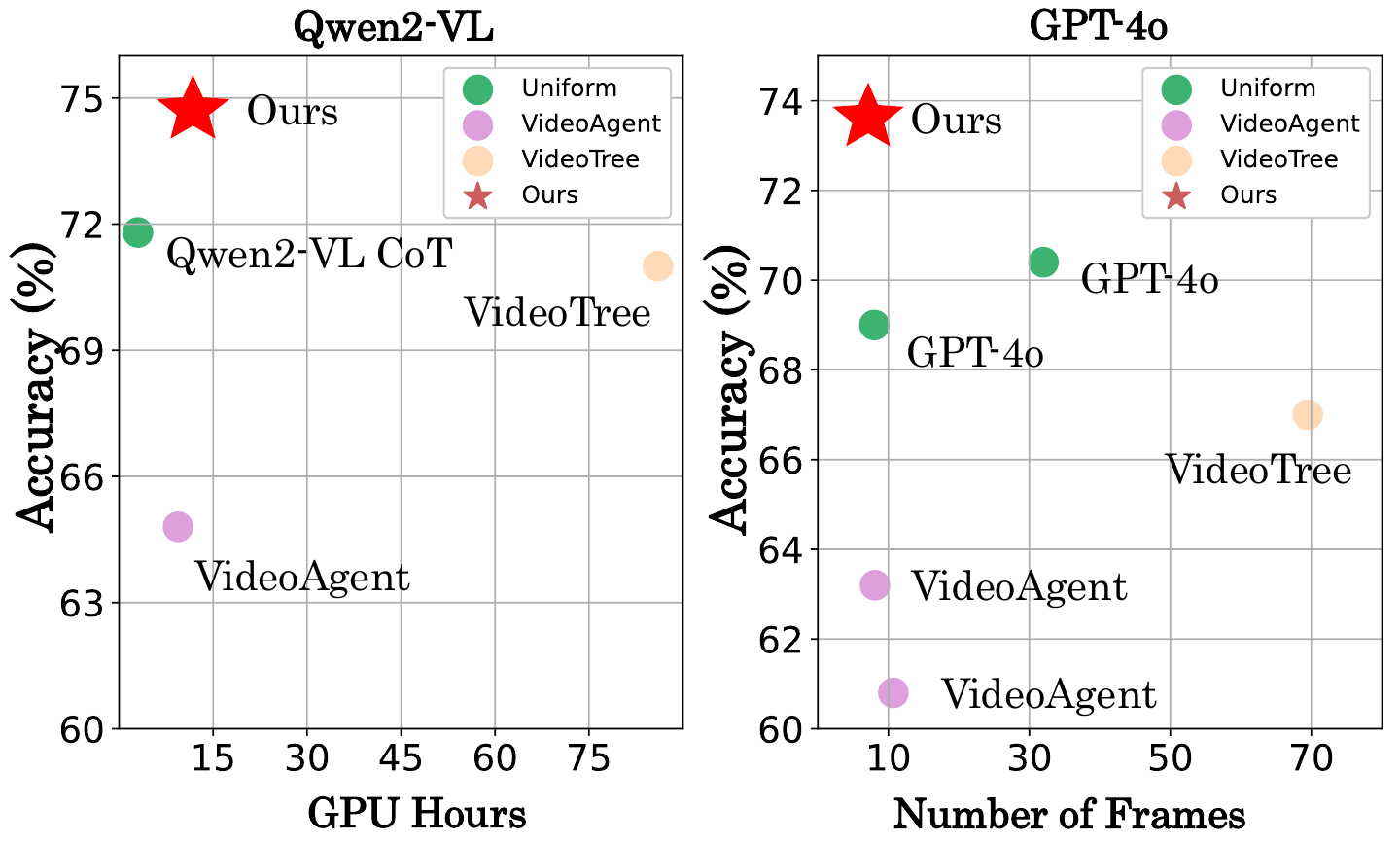}}
    \caption{\textcolor{update}{\textbf{Accuracy v.s. Computation Cost and Number of Observed Frames.} Compared to baselines, our agent demonstrates superior performance with higher efficiency (observing fewer frames and incurring comparable or lower computational cost) on EgoSchema, using both GPT-4o and Qwen2-VL-72B.}}
    \label{fig:intro}
\end{figure}

Previous works commonly utilize two-stream networks to capture spatial and temporal information~\cite{feichtenhofer2016convolutionaltwostreamnetworkfusion,Feichtenhofer_2019_ICCV} or introduce 3D operations~\cite{tran2015learningspatiotemporalfeatures3d,carreira2018quovadisactionrecognition}.
Recent approach~\cite{li2024llms,xu2023retrievalbased} leverages the long-sequence reasoning capabilities of Large Language Models (LLMs)~\cite{wang2024videollamb, achiam2023gpt} for long-form video question answering, often by converting videos into densely and uniformly sampled frames~\cite{team2024gemini,liu2024world,li2025llama}. However, this approach is usually inefficient due to the inherent imbalance and varying information density across video segments~\cite{wang2024lvbench,yu2024self}. Some segments are rich in critical details necessary for answering questions, while others contain only trivial information~\cite{wang2024videotreeadaptivetreebasedvideo}. Uniform sampling~\cite{song2023moviechat,wang2024qwen2}, which might focus on redundant frames that increase the computational burden and may overlook frames with essential information, often fails to efficiently utilize the video’s dynamic structure for question answering~\cite{VideoAgent,fan2025videoagent}.

\begin{figure}[t]
    \centering
    \includegraphics[width=\linewidth]{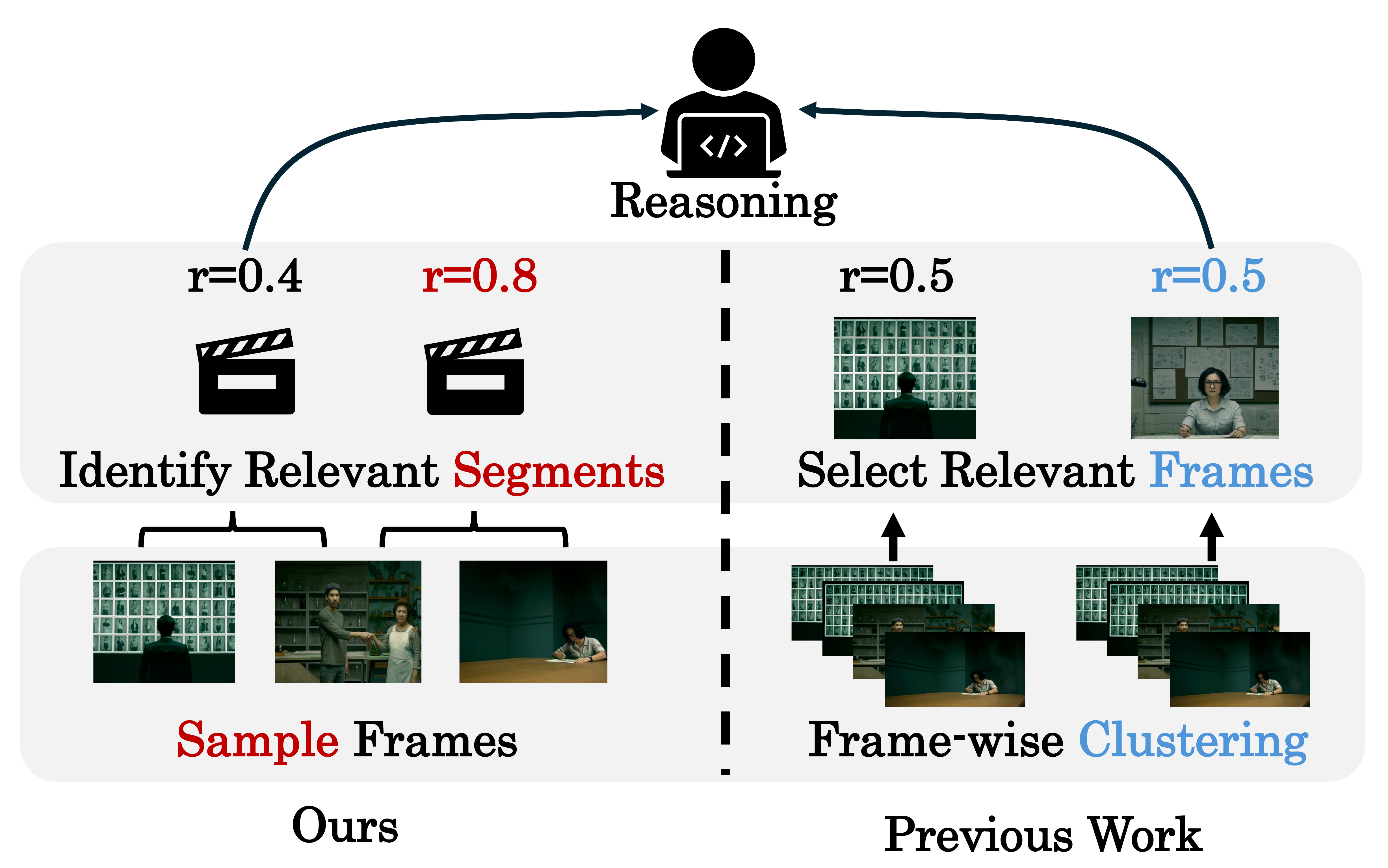}
    \caption{\textcolor{update}{\textbf{Comparison of VCA with prior works.} \textbf{Difference-1:} When answering a query, VCA uniformly samples (a few) frames, whereas previous methods (e.g., VideoTree) require captioning each frame for subsequent image retrieval, which can be computationally expensive for long videos. \textbf{Difference-2:} previous methods rely on image similarity to locate key frames, potentially losing important video features. In contrast, VCA uses segment-level exploration to identify the most informative segments.}}
    \label{fig:framework-compare}
\end{figure}

To address this limitation, recent approaches~\cite{lin2024vila, lin2023video,korbar2025text,weng2025longvlm} employ complex architectures~\cite{zhaovideoprism,wang2024internvideo2} or specialized preprocessing techniques~\cite{wang2024lifelongmemory,park2024too} to improve the efficiency of long video content processing. 
These methods~\cite{VideoAgent,wang2024videotreeadaptivetreebasedvideo} often convert videos into text by captioning densely sampled frames or incorporating memory modules~\cite{balazevic2024memory,fan2025videoagent, ma2024drvideo} to store essential information. However, these approaches still incur significant computational costs. 
\textcolor{update}{For example, captioning an hour-long video with Qwen2-VL-72B at just 1 fps requires over 3 hours on a single H100.
Additionally, these methods~\cite{chung2023long} identify key frames based on image similarity rather than a holistic understanding of the video content, which can result in missing relevant video features.}
This necessitates the need for an efficient and scalable solution for long-form video understanding.

In this work, we introduce \Model, a curiosity-driven agent inspired by how humans tackle the long video understanding task. Given a question, human is usually intrinsically motivated to identify relevant information in the video without external knowledge. 
\textcolor{update}{In practise, human tends to build a global understanding of the video, then curiously zoom in to explore specific sections in greater detail.}
\textcolor{update}{Motivated by this process, we propose a novel video agent framework, dubbed as ``\Model'', to emulate such behavior.} \Model treats the video as an environment to explore, actively navigating its contents and identifying segments that merit deeper analysis. \Model incorporates three essential techniques: (1) Tree-search Exploration: Rather than selecting specific frames~\cite{VideoAgent}, \Model uses a tree-search structure to adaptively explore the video segments in a structured way. (2) Reward Model: \Model leverages an intrinsic reward scoring mechanism to guide exploration toward the segments most relevant to the query. (3) Memory Management: \Model employs a fixed memory buffer to store key information and discard irrelevant frames, effectively reducing computational costs and avoiding memory overflow. The key difference between \Model and prior works is outlined in Fig.~\ref{fig:framework-compare}. Extensive experiments and superior performance across multiple benchmarks demonstrate that our proposed \Model framework significantly improves an agent's effectiveness and efficiency in reasoning over extended video content. More concretely, our contributions are threefold,
\begin{itemize}
    \item \Model is a long video understanding agent that can self-explore the most informative video segments based on its intrinsic reward, without relying on any auxiliary tools. 
    \item \Model is easy to implement, training-free, and is flexible to integrate with any vision-language model.
    \item As highlighted in Fig.~\ref{fig:intro}, \Model outperforms existing baselines while observing fewer frames and incurring comparable or lower computational cost, demonstrating both effectiveness and efficiency.
\end{itemize}


%% file: sec/2_related.tex
\section{Related Work}

\subsection{Long-context Multi-modal Agents}
Large multi-modal models (LMMs)~\cite{lin2023vila,agrawal2024pixtral,team2024chameleon,bavishi2023fuyu,chen2023pali} have demonstrated promising performance in addressing complex multimodal tasks~\cite{zhang2023movqa,fu2023mme,song2023moviechat,rawal2024cinepile}. However, these LMMs typically lack self-directed exploration capabilities required for handling more tricky tasks~\cite{xie2024osworld,hong2024cogagent}. To bridge this gap, multimodal agents are designed to enhance LMMs with the ability to autonomously make decisions and execute tasks within specific environments~\cite{durante2024agentaisurveyinghorizons,xie2024openagents}, such as interpreting visual cues and text prompts to make real-time decisions within websites~\cite{koh2024visualwebarenaevaluatingmultimodalagents,yao2023webshopscalablerealworldweb}.

A key feature of multimodal agents is their long-context capability~\cite{zhao2024longagent}, i.e., the ability to process and utilize information across extended periods to make informed decisions~\cite{jang2024videowebarenaevaluatinglongcontext}. An example of this is vision-language model (VLM) agents for video understanding tasks. Despite their advancements, even sophisticated VLM agents struggle to extract crucial information and perform effective reasoning over lengthy inputs~\cite{zhang2023simple,VideoAgent}. Therefore, efficiently extracting essential information and making informed decisions remains a primary challenge. This work aims to address this challenge by enabling agents to self-explore the environment and self-gather the most relevant information to efficiently solve video-understanding tasks.

\begin{figure*}[t]
    \centering
    \includegraphics[width=0.88\linewidth]{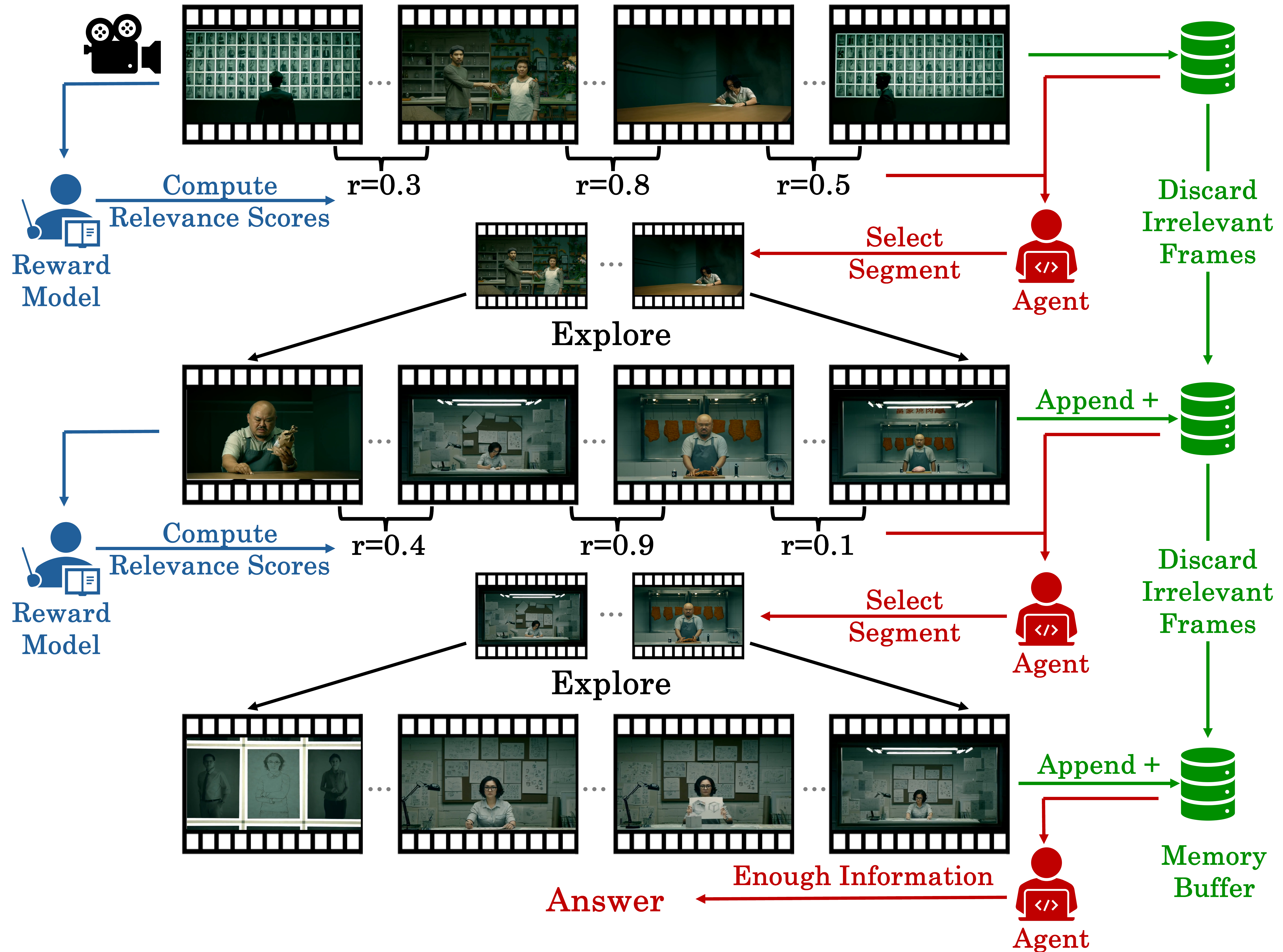}
    \caption{\textbf{Pipeline of \Model Framework.} First, the reward model predicts the relevance scores of each segment. Next, the memory buffer is updated by adding new frames and discarding irrelevant ones. Finally, emulating human reasoning, the agent actively selects segments to explore until the correct answer is found.}
    \label{fig:framework}
\end{figure*}

\subsection{Long Video Understanding}
Early deep learning models for video understanding primarily followed two main approaches~\cite{tang2024videounderstandinglargelanguage}. One uses two-stream networks~\cite{feichtenhofer2016convolutionaltwostreamnetworkfusion,Feichtenhofer_2019_ICCV,wang2016temporalsegmentnetworksgood,ng2015shortsnippetsdeepnetworks} to integrate spatial and motion information, while the second focuses on 3D CNNs~\cite{tran2015learningspatiotemporalfeatures3d,carreira2018quovadisactionrecognition,xie2018rethinkingspatiotemporalfeaturelearning} for spatial-temporal feature extraction. Vision Transformers~\cite{dosovitskiy2021imageworth16x16words} then introduce self-attention mechanisms to handle temporal modeling more effectively~\cite{9710415,fan2021multiscalevisiontransformers,bertasius2021spacetimeattentionneedvideo}. Methods with self-supervised pre-training~\cite{sun2019videobertjointmodelvideo, zhu2020actbertlearninggloballocalvideotext, tong2022videomaemaskedautoencodersdataefficient}
exhibit transferability across different tasks and achieve more robust video understanding performance.

With the development of LLM, multimodal LLM~\cite{hong2024cogvlm2visuallanguagemodels,Qwen2VL,zhang2024llavanextvideo} shows strong capabilities in video understanding tasks, but efficiently processing information within a prolonged video environment is the key challenge. Recent approaches 
often use frame sampling~\cite{yu2023self,ren2023timechat,li2024llms,xu2023retrievalbased,wang2024internvideo2} and feature extraction~\cite{sun2022long,wu2022memvit,song2023moviechat,jin2023chatunivi} to reduce computational costs. Frame sampling selects a subset of frames, which risks missing essential context, especially in lengthy videos where transitions and subtle cues may easily be neglected. Additionally, processing entire video sequences as tokens presents significant memory and computational challenges~\cite{papalampidi2024simple,wang2024videollamb,zhang2023video,cheng2024videollama}.

Recent work has explored auxiliary techniques and modules to improve efficiency, such as pre-extracted content like captions~\cite{VideoAgent,park2024too}, multimodal summaries~\cite{ma2024drvideodocumentretrievalbased,wang2024lifelongmemory}, and combinations of various auxiliary tools~\cite{fan2025videoagent,kumar2024mmctagentmultimodalcriticalthinking}. However, these tools introduce additional computational and memory costs. Moreover, some approaches still require traversing the entire video content~\cite{fan2025videoagent}. \textcolor{update}{Notably, VideoTree~\cite{wang2024videotreeadaptivetreebasedvideo} requires captioning every frame in the video, making it prohibitively expensive.} In contrast, our \Model can solve the task using a single VLM, without relying on any supplementary tools.

%% file: sec/3_approach.tex
\section{Method}
\label{sec:approach}
Inspired by human reasoning, we introduce a self-exploration agent that adaptively navigates video environments. Unlike approaches that depend on dense frame sampling or captioning~\cite{VideoAgent}, which require VLMs to process numerous frames, our agent analyzes videos iteratively within a limited context window in a process we call self-exploration. This approach allows essential information to be extracted from long videos in a coarse-to-fine manner.

Specifically, we propose a segment-based tree-search exploration structure. Given a video $v$ and a user query $q$, the agent $\pi$ begins by randomly sampling $N$ frames $\left\{f^{i}_{v}\right\}_{i=1}^{N}$ to obtain an initial overview of the content in each video segment $v_i \in \left\{v_{i}\right\}_{i=1}^{N+1}$. The agent then decides whether additional exploration is needed to answer the query. If further investigation is necessary, the agent selects a specific segment $s^{*} \in \left\{v_{i}\right\}_{i=1}^{N+1}$ for deeper exploration until sufficient information is gathered to respond to the query. 

Nevertheless, a one-hour movie typically contains approximately 86,400 frames, posing a challenge for current VLM-based agents to efficiently identify the most relevant segment~\cite{VideoAgent,fan2025videoagent,wang2024videotreeadaptivetreebasedvideo}. Moreover, even with a small $N$, conducting multiple rounds of exploration accumulates a substantial number of frames, leading to high computational costs. To address these issues, we introduce a reward model $R$ to guide the agent's exploration and propose a memory buffer $M$ to reduce storage and inference costs.

In this section, we first explain how the reward model is integrated and assists the agent throughout the exploration process (Sec.~\ref{sec:reward-model}). Next, we introduce our memory management strategy in Sec.~\ref{sec:memory-management}. Finally, we explain how \Model adaptively explores the video using a segment-based tree structure (Sec.~\ref{sec:tree-search-exploration}).

\subsection{Reward Model}
\label{sec:reward-model}
With the segment-based tree-search exploration, the agent is required to select the most informative video segment for exploration at each step. However, we observe that current VLM-based agents struggle to independently identify optimal exploration paths (see Sec.~\ref{sec:exp-ablation}). To this end, inspired by the selective attention mechanism in human cognition~\cite{desimone1995neural}, we incorporate a reward model $R$ to guide the agent’s selection process, i.e., the agent will gain a higher reward if it plans to take the action to explore more relevant segment.

Specifically, during the exploration of a current selected segment $s^*$, the reward model observes a set of frames $\left\{f^{i}_{s^*}\right\}_{i=1}^{N}$ sampled from $s^*$. These frames $\left\{f^{i}_{s^{*}}\right\}_{i=1}^{N}$ then form finer sub-segments $\left\{s^{*}_i = \lbrack l_{s^{*}}^i, l_{s^{*}}^{i+1}\rbrack\right\}_{i=1}^{N+1}$, where $l_{s^{*}}^i$ and $l_{s^{*}}^{i+1}$ represent the indices of the beginning and end frame of the sub-segment, respectively. The reward model aims to generate a reward score for each sub-segment $\left\{s^{*}_i\right\}_{i=1}^{N+1}$  based on its relevance to the user query $q$. We utilize chain-of-thoughts~\cite{wei2022chain} technique to let the reward model provide a verbal explanation of each segment and then generate a relevant score, i.e., $\left\{t_{s^*_{i}}; r_{s^*_{i}}\right\}_{i=1}^{N+1}$. Additionally, to maintain consistency across different exploration steps, we incorporate the reward history $H_r$ from previous steps alongside the new segments during evaluation. A simplified prompt template for the reward model is presented here, with detailed prompts in Appendix~\ref{app:prompt-design}.

\begin{tcolorbox}[title=Reward Model Prompt (Simplified)]
    Given: \\
    $H_r$, reward score history in previous steps. \\
    $\left\{f^{i}_{s^*}\right\}_{i=1}^{N}$, newly sampled frames from current segment $s^{*}$. \\
    Please first explain each sub-segment $\left\{s^{*}_i\right\}_{i=1}^{N+1}$ and then assign a relevance score of each sub-segment to the user query $q$.
\end{tcolorbox}

In practice, rather than training a separate specialized model, we utilize a same VLM-based agent to function as the reward model. While the reward model $R$ and exploration agent $\pi$ serve distinct purposes, both can be implemented using the same model. In this work, we employ a single VLM-based agent for both roles across different settings, providing a neat and efficient agent framework that is guided by its self-generated intrinsic reward and without relying on auxiliary models or tools.
\subsection{Memory Management}
\label{sec:memory-management}
As exploration proceeds over multiple rounds, the accumulation of frames can lead to significant computational costs for long videos, even when $N$ is small. However, as illustrated in Fig.~\ref{fig:framework-compare}, 
they tend to gradually disregard irrelevant frames, especially from earlier steps, retaining only the most crucial information in their working memory for later analysis. Inspired by this, we introduce a fixed-size memory buffer $M$ that stores only the most relevant frames throughout exploration.

Specifically, when the agent $\pi$ explores a video segment $s^{*}$, the associated frames $\left\{f^{i}_{s^*}\right\}_{i=1}^{N}$ are added to the memory buffer. If the total number of frames exceeds the buffer limit, irrelevant frames are removed based on relevance scores $r_{s^*_i}$ defined in Sec.~\ref{sec:reward-model}. We discard the frames with the lowest scores and only retain the most relevant ones. While these less relevant frames are removed, their visual information is extracted and encoded as text descriptions $\left\{t_{s^*_{i}}\right\}_{i=1}^N$ within the exploration history $H_r$, allowing the reward model to utilize. This memory management strategy enables the agent to conduct reasoning within a fixed resource limit, regardless of the exploration trajectory's length, thereby preventing excessive memory usage.
\subsection{Tree-Search Exploration}
\label{sec:tree-search-exploration}
\textcolor{update}{Inspired by humans adaptively focusing on different video segments during exploration rather than targeting specific frames directly,} our proposed agent \Model treats the video as an unexplored environment, where each segment represents a node on an exploration tree. Given the entire video as the root node, the agent’s goal is to efficiently identify an optimal short path leading to a leaf node (segment) containing the most informative content needed to answer the query.

At each step, the agent $\pi$ receives both textual and visual guidance: a set of candidate segments $S$, which includes a collection of segments with associated reward scores, and a memory buffer $M$ containing multiple frames. If the agent has gathered sufficient information to answer the user’s query $q$, it directly generates a response $a$. Otherwise, it continues exploring by selecting a specific segment $s^* \in S$ to expand the tree based on reward scores. 

However, due to the limited capacity of current foundational models, these reward scores are not always accurate (see further analysis in Sec.~\ref{sec:reward-quality}). Selecting segments based on reward scores in a purely greedy manner can potentially lead to a suboptimal solution. Instead, we use the reward scores as guidance for agent $\pi$, allowing it to make the final segment selection independently. In this case, the agent is not forced to select the segment with the highest reward, balancing the exploitation and exploration. More importantly, the candidate set $S$ includes both unexplored coarse segments from prior steps and fine-grained segments from the current step, enabling the agent to backtrack and select earlier segments if it detects that the current exploration path may be suboptimal. Below are the simplified prompts. The full prompt is provided in Appendix~\ref{app:prompt-design}.
\begin{tcolorbox}[title=Exploration Agent Prompt (Simplified)]
    Given: \\
    Candidate segments set: $S$ \\
    Frames in Memory buffer: $M$ \\
    Please choose one segment to explore based on the query $q$. Output answer if you have gathered sufficient information.
\end{tcolorbox}

The overall framework of our proposed method is summarized in Algo.~\ref{algo:self-exploration}. Leveraging a tree-search exploration technique boosted with a reward model and a memory buffer, our self-exploration agent adaptively and efficiently explores the video like a human. Empirical results in Sec.~\ref{sec:exp-ablation} further validate the effectiveness of our tree-search exploration over naive frame selection methods.


\begin{algorithm}
\caption{\Model: Video Curious Agent}
\label{algo:self-exploration}
\begin{algorithmic}[1]
\REQUIRE Video environment $v$, question $q$, agent $\pi$, reward model $R$, memory buffer $M$, candidate segments set $S$, reward score history $H_r$, sampling frame number $N$.
\ENSURE Answer $a$
\STATE Initialize selected segment $s^{*} \gets v$
\STATE Initialize candidate segments set $S \gets \emptyset$
\STATE Initialize memory buffer $M \gets \emptyset$
\STATE Initialize reward score history $H_r \gets \emptyset$
\STATE \texttt{continue} $\gets \TRUE$
\WHILE{\texttt{continue}}
    \STATE \texttt{\textcolor{blue}{\# Sampling frames \& form segments}}
    \STATE $\left\{f^{i}_{s^*}\right\}_{i=1}^{N}, \left\{s^*_{i}\right\}_{i=1}^{N+1} \gets \text{UniformSample}(s^{*})$
    \STATE \texttt{\textcolor{blue}{\# Scoring each segment}}
    \STATE $\left\{t_{s_i^*}; 
    r_{s_i^*}\right\}_{i=1}^{N+1}\gets R(\left\{s^*_{i}\right\}_{i=1}^{N+1}; q, \left\{f^{i}_{s^*}\right\}_{i=1}^{N}, H_r)$
    \STATE \texttt{\textcolor{blue}{\# Update candidate segment set}}
    \STATE $S \gets S \cup \left\{s^*_{i}; r_{s^*_{i}}\right\}_{i=1}^{N+1} $
    \STATE \texttt{\textcolor{blue}{\# Update history with thoughts}}
    \STATE $H_r \gets H_r \cup \left\{t_{s_i^*}; r_{s_i^*}\right\}_{i=1}^{N+1}$
    \STATE \texttt{\textcolor{blue}{\# Update memory with frames}}
    \STATE $M \gets \text{UpdateMemory}(M, \left\{f^{i}_{s^*}\right\}_{i=1}^{N}; S)$
    \STATE \texttt{\textcolor{blue}{\# Output answer or select segment}}
    \STATE $a, s \gets \pi( q, S; M)$
    \IF{$a = \emptyset$}
        \STATE \texttt{\textcolor{blue}{\# Continue exploration}}
        \STATE $s^{*} \gets s$ 
    \ELSE
        \STATE \texttt{continue} $\gets \FALSE$
    \ENDIF
\ENDWHILE
\RETURN $a$
\end{algorithmic}
\end{algorithm}


%% file: sec/4_experiments.tex
\section{Experiments}

\subsection{Experimental Settings}
\label{sec:exp-setup}

\paragraph{Benchmark.} We mainly evaluate \Model on two long-term video question-answering benchmarks.
\textbf{EgoSchema}~\cite{mangalam2023egoschema} is an egocentric video dataset sourced from Ego4D~\cite{grauman2022ego4d}, with a duration of 180 seconds.
Here we use the validation set of 500 QA pairs.
\textbf{LVBench}~\cite{wang2024lvbench} covers about 1,500 samples for evaluation of reasoning abilities from high-quality videos over 30 minutes in length. We also conduct experiments on other relevant benchmark datasets in Appendix~\ref{app:experiments}.






\paragraph{Baselines.}
We compare with most relevant baselines to our work, including state-of-the-art open-source Video-LLMs~\cite{liu2024llavanext, ren2024timechat, chen2024far, hong2024cogvlm2, zhang2024longva, Qwen2VL}, the close-source model GPT-4o~\cite{hurst2024gpt}, and agent-based systems~\cite{VideoAgent, wang2024videotreeadaptivetreebasedvideo,park2024too}.
For video models, we report the results obtained with uniform sampling frames, while for the agent-based systems, we follow their official implementation with GPT-4o~\cite{hurst2024gpt}, consistent with \Model's setup to ensure a fair comparison.


\paragraph{Implementation Details.}
In this work, we use the same model for the exploration agent and the reward model.
Unless otherwise stated, all experiments were performed using the August 2024 version of \texttt{gpt-4o} as the base model, leveraging the OpenAI LLM API service\footnote{\url{https://platform.openai.com/docs/models}} with a temperature of 0.5. We set memory buffer sizes set to 8 and 16 frames for EgoSchema and LVBench, respectively. Further implementation details are provided in Appendix~\ref{app:implementation-details}.



\begin{table*}[t]
    \centering
    \begin{tabular}{@{}lcrrrrrrr@{}}
        \toprule
        Method & Frames & ER & EU & KIR & TG & Rea & Sum & Avg.\\
        \midrule
        TimeChat$^*$~\cite{ren2024timechat} & $>$ 96 & 21.9 & 21.7 & 25.9 & 22.7 & 25.0 & 24.1 & 22.3 \\
        LLaVA-NeXT$^*$~\cite{zhang2024llavanextvideo} & 32 & 30.1 & 31.2 & 34.1 & 31.4 & 35.0 & 27.6 & 32.2 \\
        InternVL2-40B$^*$~\cite{chen2024far} & 16 & 37.4 & 39.7 & \textbf{43.4} & 31.4 & 42.5 & \textbf{41.4} & 39.8 \\
        GLM4V-Plus$^*$~\cite{hong2024cogvlm2} & 20 & 39.9 & 35.8 & 34.8 & 37.7 & 40.0 & 32.8 & 38.3 \\
        \midrule
        GPT-4o$^*$~\cite{hurst2024gpt} & 64 & 35.9 & 30.8 & 35.5 & 28.3 & 33.5 & 34.5 & 34.7 \\
        VideoAgent$^\dagger$~\cite{VideoAgent} & Avg. 25.5 &28.0 &30.3 &28.0 &29.3 &28.0 &36.4 &29.3\\
        VideoTree$^\dagger$~\cite{wang2024videotreeadaptivetreebasedvideo} & Avg. 103.2 &30.3 &25.1 &26.5 &27.7 &31.9 &25.5 &28.8\\
        \midrule        \Model (ours) & Avg. 20.0 & \textbf{43.7} & \textbf{40.7} & 37.8 & \textbf{38.0} & \textbf{46.2} & 27.3 & \textbf{41.3} \\
        \bottomrule
    \end{tabular}
    \caption{\textbf{Experimental Results on LVBench}. Results of methods marked with $^*$ are sourced from LVBench~\cite{wang2024lvbench}. Methods marked with $^\dagger$ indicate implementations where \textbf{GPT-4o} serves as the main component. Our agent's memory buffer size is set to \textbf{16 frames}. We report the average number of observed frames for agent-based methods, with the best performance highlighted in \textbf{bold}.}
    \label{exp:lvbench-main}
\end{table*}

\begin{table}[t]
    \centering
    \begin{tabular}{@{}lcr@{}}
        \toprule
        Method & Frames & Subset \\
        \midrule
        \multirow{2}{*}{GPT-4o~\cite{hurst2024gpt}} & 8 & 69.0 \\
         & 32 & 70.4 \\
        \midrule
        \multirow{2}{*}{VideoAgent$^\dagger$~\cite{VideoAgent}} & Avg. 8.1 & 63.2 \\
         & Avg. 10.7 & 60.8 \\
        VideoTree$^\dagger$~\cite{wang2024videotreeadaptivetreebasedvideo} & Avg. 69.5 & 67.0 \\
         LVNet$^\dagger$~\cite{park2024too} & 12 & 68.2  \\
        \midrule
        \Model (ours) & Avg. 7.2 & \textbf{73.6} \\
        \bottomrule
    \end{tabular}
    \caption{\textbf{Experimental Results on EgoSchema}. Methods marked with $^\dagger$ indicate implementations where \textbf{GPT-4o} is the main component. The results of LVNet are from~\cite{park2024too}. Our agent's memory buffer size is set to \textbf{8 frames}. We report the average number of observed frames, with the best performance highlighted in \textbf{bold}.}
    \label{exp:egoschema-main}
\end{table}

\subsection{Experimental Results}
\label{sec:exp-result}

The experimental results on LVBench and EgoSchema are presented in Tab.~\ref{exp:egoschema-main} and Tab.~\ref{exp:lvbench-main}, respectively.
By adaptively focusing on critical segments rather than scanning the entire video, our agent not only achieves stronger performance but also incurs less computational cost. With less than 30\% of the frames, \Model achieves substantial improvements over directly feeding uniformly sampled frames into GPT-4o. Specifically, we observe performance gains of 3.2\% on the EgoSchema subset and 6.6\% on LVBench.
Significant increases in the \emph{Event Understanding} and \emph{Temporal Grounding} domains further underscore our effectiveness in identifying crucial segments within long videos.

Furthermore, we compare our framework with recently proposed SOTA video agent systems, VideoAgent~\cite{VideoAgent} and VideoTree~\cite{wang2024videotreeadaptivetreebasedvideo}. For a fair comparison, we re-implement both frameworks using GPT-4o for all reasoning components, consistent with our setup.
\textcolor{update}{However, captioning all frames with GPT-4o is prohibitively expensive. Hence, following the baseline implementations, we use open-source VLMs for captioning. A comparison using the same model for all components is provided in Sec.~\ref{sec:open-source}.}
As shown in Tab.~\ref{exp:lvbench-main}, \Model outperforms all baselines over 5\% on EgoSchema, while requiring processing fewer frames. These results suggest the limitations of dense captioning for video preprocessing tasks, which can incur more computational costs while introducing more redundant information. In contrast, our proposed memory management strategy effectively improves efficiency by only retaining the most crucial information for reasoning. Similar insights can be drawn from results on LVBench in Tab.~\ref{exp:egoschema-main}. Additional results are provided in Appendix~\ref{app:experiments}.

Moreover, we observe that \Model explores approximately three times more frames on LVBench than on EgoSchema. Recall that EgoSchema videos have an average duration of 180 seconds, while videos in LVBench extend to at least 30 minutes. This phenomenon suggests that our proposed tree-search exploration technique actively explores the video to locate the most relevant information and adaptively increases the exploration budget based on task complexity. For example, \Model exhibits both exploitation and exploration: when the agent determines that the current exploration path is suboptimal, it backtracks to previous segments to explore a new path (see Sec.~\ref{sec:case-study}). These empirical observations further validate the robustness and scalability of our approach, especially in diverse long-video contexts.

\subsection{Ablation Study}
\label{sec:exp-ablation}

In this section, we conduct an ablation study to evaluate the components of our agent.
Results across both benchmarks are presented in Tab.~\ref{exp:ablation-egoschema-lvbench}. As illustrated in the table, performance consistently declines when the reward model is removed, highlighting that the agent’s inherent reasoning ability alone is insufficient for accurately identifying the most informative segments. This finding emphasizes the effectiveness of incorporating an intrinsic reward to guide the agent's segment selection process. A detailed analysis of our reward model's performance is provided in Sec.~\ref{sec:reward-quality}.

\begin{table}[!t]
    \centering
    \scalebox{1}{
    \begin{tabular}{@{}lcrcr@{}}
        \toprule
        \multirow{2}{*}{Method} & \multicolumn{2}{c}{EgoSchema} & \multicolumn{2}{c}{LVBench} \\
        & Frames & Acc. & Frames & Acc. \\
        \midrule
        \Model (ours) & 7.2 & \textbf{73.6} & 20.0 & \textbf{41.3} \\
        - w/o Reward & 7.5 & 72.6 & 22.2 & 39.5 \\
        - w/o Tree Search & 10.2 & 69.8 & 39.9 & 36.2 \\
        \bottomrule
    \end{tabular}
    }
    \caption{\textbf{Ablation Study on EgoSchema and LVBench}. Both the tree-search exploration and reward model contribute significantly to improvements in performance and efficiency. Detailed ablation study results on LVBench are included in Appendix~\ref{app:experiments}.}
    \label{exp:ablation-egoschema-lvbench}
\end{table}

\begin{table}[t]
    \centering
    \scalebox{0.95}{
    \begin{tabular}{@{}lrrr@{}}
        \toprule
        Buffer Size &8 &16&32 \\
        \midrule
        Acc.  & 38.9 & 41.3 & \textbf{43.2} \\
        \bottomrule
    \end{tabular}
    }
    \caption{\textcolor{update}{\textbf{Ablation Study of Buffer Size on LVBench.} Increasing the buffer size allows the model to observe more frames during reasoning, leading to improved performance.}}
    \label{exp:ablation-buffer-size}
\end{table}

Additionally, we observe a significant performance drop when the tree-search exploration technique is ablated, i.e., the agent directly selects frame indices without the segment-based tree structure. Interestingly, this modification also results in a substantial increase in the average number of frames acquired, especially on LVBench. This indicates that the agent struggles to locate the most crucial information and consequently produces a longer but redundant trajectory. These ablation study results validate the effectiveness of the tree-search exploration framework assisted by the reward model. Additionally, this underscores \Model’s potential to achieve greater efficiency and improved performance with more advanced VLMs and reward models.


\textcolor{update}{
Moreover, as shown in Tab.~\ref{exp:ablation-buffer-size}, increasing the memory buffer size yields consistently improved performance for VCA, demonstrating that our framework benefits from a broader reasoning window while maintaining effectiveness with fewer frames. Due to limited computation resources, we only tested up to 32 frames; however, the upward trend indicates the potential for gains with additional frames.
}

\begin{figure*}[!t]
\centering
\centerline{\includegraphics[width=0.8\linewidth]{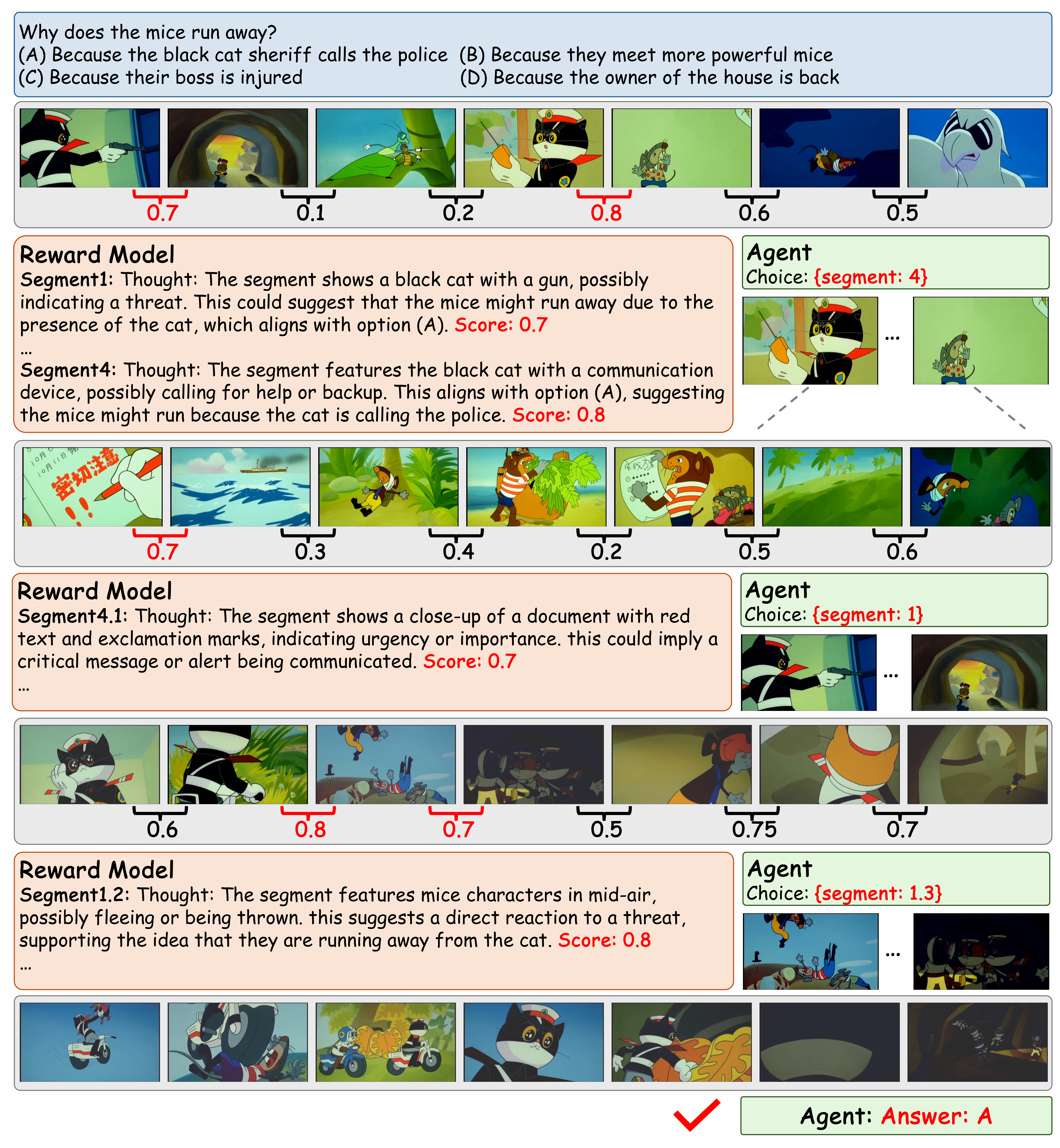}}
\caption{\textbf{Exploration Trajectory Example.} The agent generally tends to select video segments with higher reward scores for exploration, yet it can adaptively redirect to earlier segments, effectively balancing exploitation and exploration.}
\label{fig:good-case}
\end{figure*}

%% file: sec/5_analysis.tex
\section{Analysis}

\subsection{Can \Model generalize to Open-source Models?}
\label{sec:open-source}

While \Model demonstrates strong performance using GPT-4o as the base model, we also analyze whether these positive results can be extrapolated to open-source models. 
\textcolor{update}{We implement the \Model framework with Qwen2-VL-72B-Instruct-GPTQ-Int4~\cite{Qwen2VL} as both the exploration agent and reward model, as well as the captioner for baselines.
The results in Tab.~\ref{exp:egoschema-local-model} reveal that, compared to directly feeding uniformly sampled frames into Qwen2-VL, our framework achieves higher accuracy on the EgoSchema benchmark despite using only about 60\% of the frames. In contrast, VideoAgent performs significantly worse. }

\textcolor{update}{Furthermore, our approach outperforms VideoTree with far fewer frames. 
Notably, VideoTree employs clustering to build a caption similarity-based tree structure for frame selection, requiring caption generation for each frame. This process is computationally expensive, demanding over 60 GPU hours on a single H100, as shown in Fig.~\ref{fig:intro}.
These results further underscore the robustness of \Model.}

\begin{table}[t]
    \centering
    \begin{tabular}{@{}lcr@{}}
        \toprule
        Method & Frames & Subset\\
        \midrule
        \multirow{2}{*}{Qwen2-VL} & 8 & 70.2 \\
        & 32 & 74.2 \\
        \midrule
        VideoAgent & Avg. 8.6 & 65.2 \\
        VideoTree & Avg. 85.4 & 71.0\\
        \midrule
        \Model (ours)  & Avg. 16.9 & \textbf{75.2} \\
        - w/ Image Reward & Avg. 20.9 & 74.0 \\
        \bottomrule
    \end{tabular}
    \caption{\textbf{Performance of Using Open-source Model on EgoSchema.} Our approach achieves higher accuracy with fewer observed frames, showing consistency improvements over baseline methods. Memory buffer size is set to 16 frames.}
    \label{exp:egoschema-local-model}
\end{table}

\subsection{What's the Exploration Behavior of \Model?}
\label{sec:case-study}
Here we present a real exploration trajectory from LVBench to illustrate our agent's decision-making process. As shown in Fig.~\ref{fig:good-case}, given a user query, the reward model initially evaluates each segment’s relevance based on uniformly sampled frames, and the agent selects the fourth segment, which has the highest relevance score. However, upon further investigation of segment 4, both the reward model and the agent find it relatively irrelevant to the query. Rather than continuing with it, the agent redirects its exploration to segment 1 in the last round. After selecting segment 1, the agent chooses to investigate the third segment, despite the second segment having the highest reward score. This behavior reflects \Model’s ability to adaptively adjust its exploration trajectory based on visual context rather than purely following the reward score, resulting in a more robust system.

We also examine the limitations of \Model by analyzing its failure cases, which fall into the following scenarios: The most common failure occurs when the agent overlooks subtle clues, resulting in an inaccurate response to the user query even after prolonged exploration, particularly for queries requiring specific visual details. Other failure cases include instances where the agent is misled by inaccurate reward scores or fails to provide the correct answer despite observing key frames. Detailed cases and further discussion are provided in the Appendix~\ref{app:case-study}. 

\textcolor{update}{In addition, unlike previous methods, \Model employs a segment-level exploration strategy instead of direct image similarity to select key frames. 
To further investigate this difference, we evaluate the performance of selecting the segment whose start and end images have the highest visual relevance. 
As shown in Tab.~\ref{exp:egoschema-local-model}, relying solely on image relevance instead of inferring segment semantics results in a performance drop on EgoSchema, confirming the effectiveness of our segment-level exploration.}

\subsection{How Reliable is the Reward Model?}
\label{sec:reward-quality}
\begin{figure}[!t]
\centering
\centerline{\includegraphics[width=\linewidth]{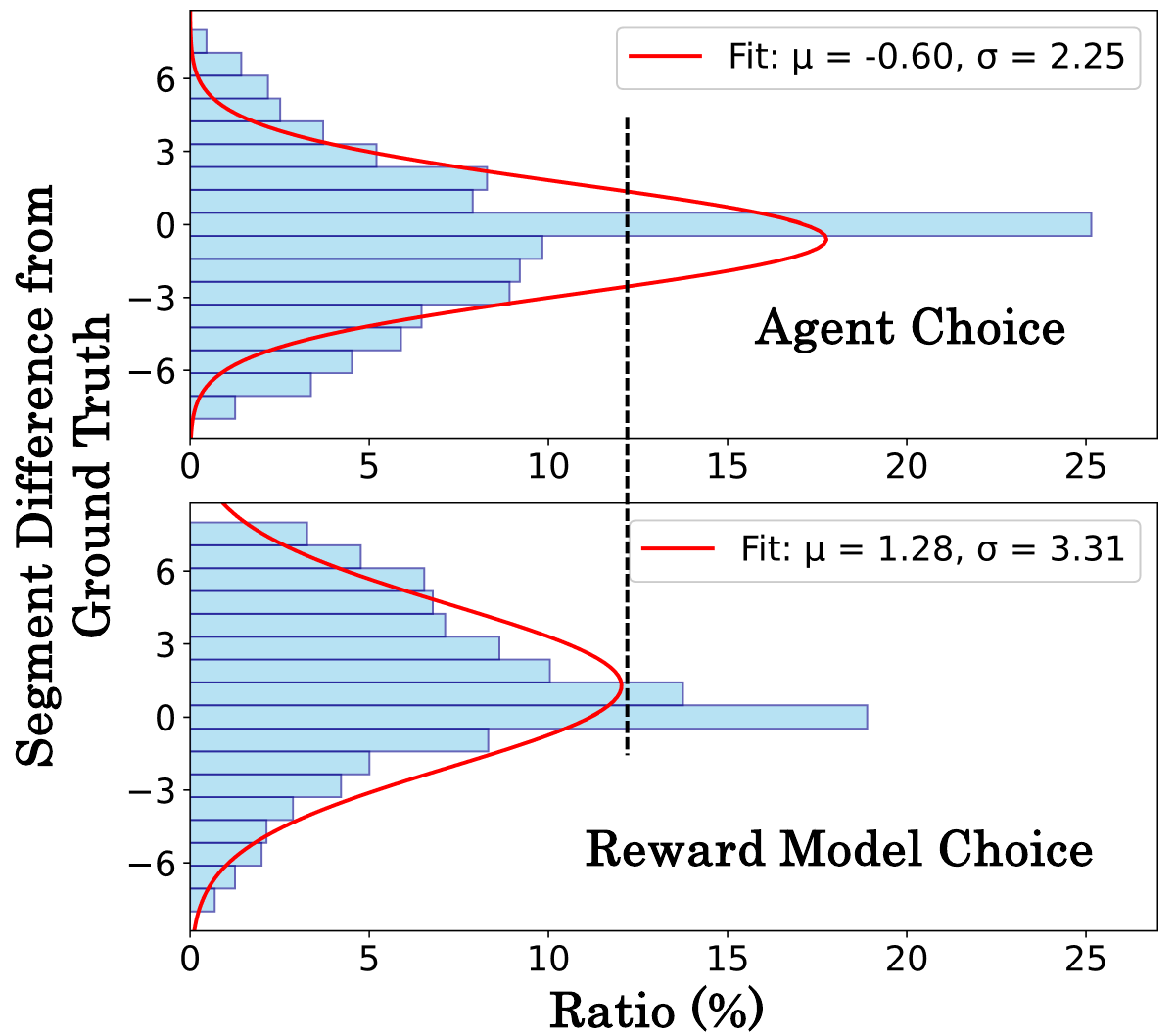}}
\caption{\textbf{Comparison of Segment Selection Accuracy between Reward-guided Agent and Reward Model on LVBench.} The y-axis represents the segment distance from the ground truth, and the x-axis represents the ratio of counts. The red curves represent fitted normal distributions. The agent's selection aligns closer to the ground truth compared to the reward model's greedy selection.}
\label{fig:reward-match}
\end{figure}

To further assess the capability of our reward model, we visualize the matching accuracy between segments with the highest reward scores and ground truth time references from LVBench. The results are shown at the bottom of Fig.~\ref{fig:reward-match}. We observe that segment distances follow an approximately normal distribution with a mean of $1.28$, indicating that the segment with the highest reward score often closely aligns with the ground truth segment. This validates the reward model’s effectiveness in providing meaningful guidance.

Additionally, as discussed in Sec~\ref{sec:tree-search-exploration}, we encourage the agent to make independent decisions when selecting segments rather than blindly following reward scores in a purely greedy manner. The benefit of this design choice is demonstrated at the top of Fig.~\ref{fig:reward-match}, where we present the matching accuracy of the agent’s choices. Compared to the greedy approach, the agent’s selections result in a sharper distribution of segment distances, with an average distance gap of $-0.60$, indicating closer alignment with the ground truth. In summary, our reward model assigns reliable relevance scores to segments, while the agent further refines these choices to achieve even more accurate selections.

\begin{table}[!t]
    \centering
    \begin{tabular}{@{}lcrcr@{}}
        \toprule
        Method & Frames & Acc. & Frames & Acc. \\
        \midrule
        & \multicolumn{2}{c}{(LongVA)} & \multicolumn{2}{c}{(GPT-4o)} \\
        \Model (ours) & 13.7 & 33.4 & 20.0 & 41.3 \\
        - w/ GT Reward & 27.2 & \textbf{36.1} & 24.0 & \textbf{44.2} \\
        \bottomrule
    \end{tabular}
    \caption{\textbf{Comparison of Accuracy w/ and w/o Ground Truth (\emph{GT}) Reward Scores on LVBench}. Incorporating \emph{GT} reward scores further improves performance on both LongVA and GPT-4o models. Memory buffer size is set to 16 frames.}
    \label{exp:ablation-ground-truth-score}
\end{table}

\subsection{Can \Model Benefit from Better Reward?}
\label{sec:reward-gt}
Previous analysis demonstrates the effectiveness of the reward model, though a gap remains compared to ground truth guidance. To examine the upper bound of our agent's performance with optimal reward guidance, we substitute the reward scores with ground truth time references. The results, shown in Tab.~\ref{exp:ablation-ground-truth-score}, indicate that using ground truth scores yields a consistent 3\% improvement on LVBench, regardless of whether GPT-4 or LongVA is used as the model. This finding highlights the great potential of our framework when paired with stronger, specialized reward models that can provide more accurate and informative guidance.


%% file: sec/6_conclusion.tex
\section{Conclusion}
In this work, we propose a curiosity-driven video agent framework for long video understanding tasks with self-exploration capability. VCA achieves state-of-the-art performance and efficiency without relying on auxiliary tools. In Sec.~\ref{sec:reward-gt}, we demonstrate VCA’s great potential with more accurate reward guidance, suggesting that collecting synthetic data and training a specialized reward model would be a valuable direction for future work.

%% file: sec/X_suppl.tex
\clearpage
\setcounter{page}{1}
\maketitlesupplementary

\section{Prompt Design}
\label{app:prompt-design}

\begin{figure*}[h]
\centering
\centerline{\includegraphics[width=0.9\linewidth]{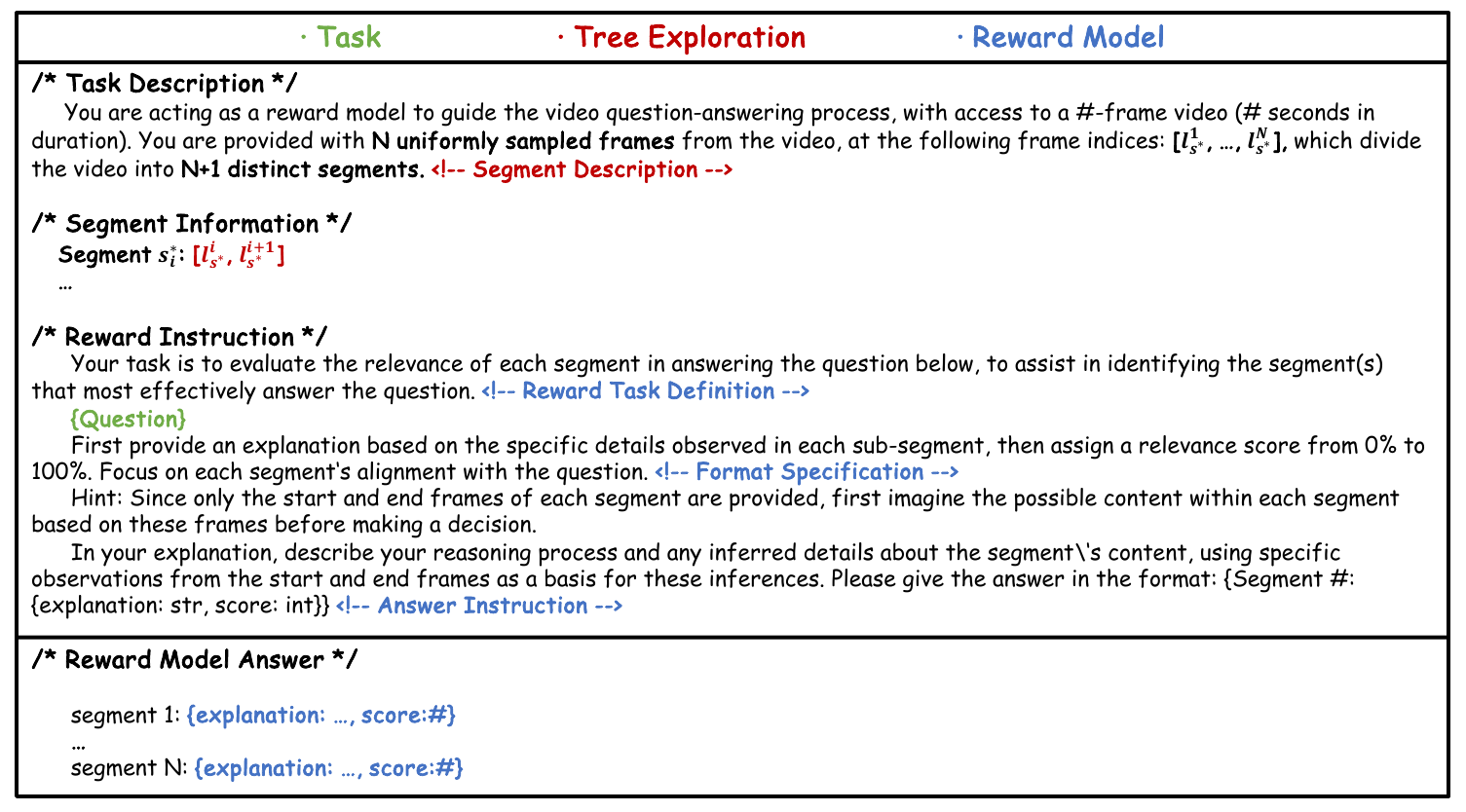}}
\caption{\textbf{Reward Model ($R$) Prompt Template for the First Round}. The prompt template begins by defining the reward scoring task and presenting the segment candidates. Given the target query, the reward model is instructed to first provide an explanation for its reasoning and then assign a relevance score to each candidate. \texttt{\textless!-- --\textgreater} represents comments or explanations about the given prompt.}
\label{fig:prompt-reward}
\end{figure*}

\begin{figure*}[h]
\centering
\centerline{\includegraphics[width=0.9\linewidth]{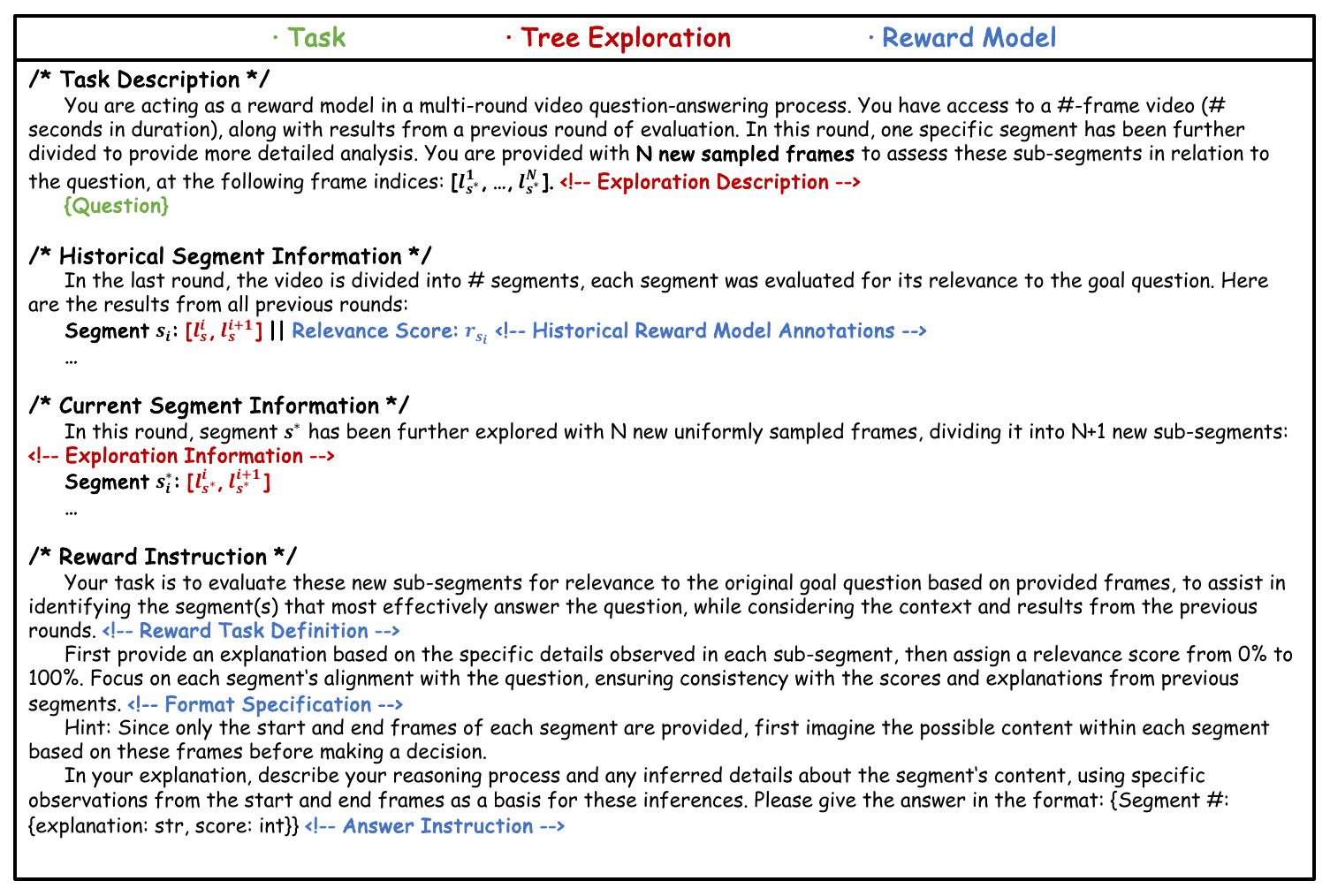}}
\caption{\textbf{Reward Model ($R$) Prompt Template for the Subsequent Rounds}. This prompt template, used after the initial round, begins by outlining the exploration strategy and presenting the target query. Besides, to ensure consistency in relevance scores throughout the exploration trajectory, it also includes historical segment information along with the reward annotations previously assigned. \texttt{\textless!-- --\textgreater} represents comments or explanations about the given prompt.}
\label{fig:prompt-reward-followup}
\end{figure*}

\begin{figure*}[h]
\centering
\centerline{\includegraphics[width=0.9\linewidth]{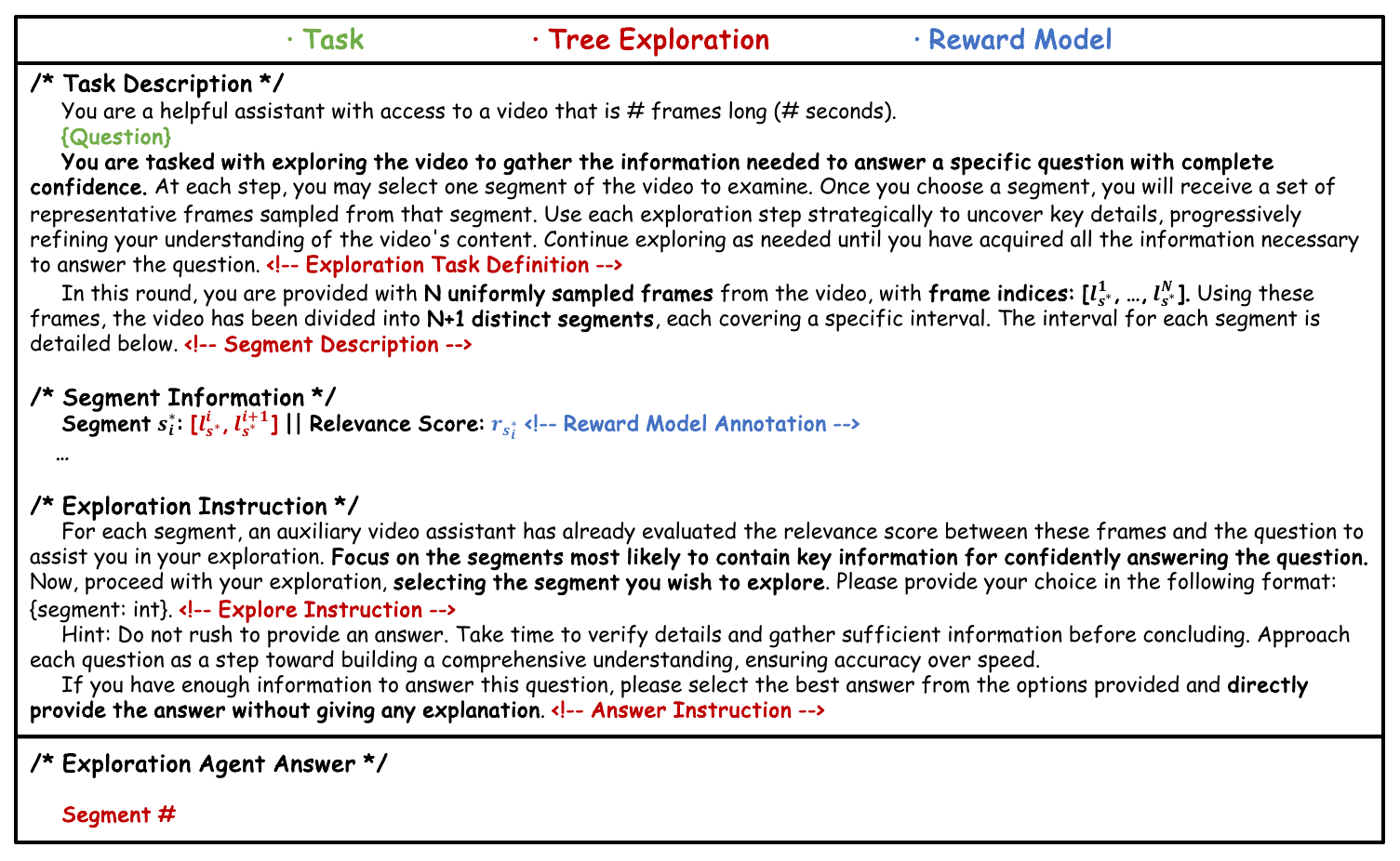}}
\caption{\textbf{Exploration Agent ($\pi$) Prompt Template}. The prompt template begins by introducing the target question, followed by a definition of the exploration task. Next, the segment candidates' information is provided, guiding the agent to either select a segment for further exploration or answer the question if sufficient information has been gathered. \texttt{\textless!-- --\textgreater} represents comments or explanations about the given prompt.}
\label{fig:prompt-exploration}
\end{figure*}

In this section, we elaborate the prompt design of our VCA framework. We include the prompt on how the reward model ($R$) generates relevance scores in Fig.~\ref{fig:prompt-reward} and Fig.~\ref{fig:prompt-reward-followup}, and the prompt used for the exploration agent ($\pi$) in Fig.~\ref{fig:prompt-exploration}, respectively. 

As illustrated in in Fig.~\ref{fig:prompt-reward}, for the first round, the reward model starts evaluating the relevance of each segment to the given question. For the subsequent rounds of exploration, since the reward model can only access the sampled frames from the selected local segment, we include reward history ($H_r$) in previous rounds to ensure consistency and alignment in terms of relevance score scale. The detailed template is shown in Fig.~\ref{fig:prompt-reward-followup}.

As illustrated in in Fig.~\ref{fig:prompt-exploration}, the Exploration Agent determines the next action based on these relevance scores. If the available information is sufficient, the agent proceeds to provide an answer. Otherwise, it continues to iteratively refine its exploration. Specifically, for the exploration agent, all frames stored in the memory buffer are presented together with the dialog history that is formatted in a conversational style.

\section{Implementation Details}
\label{app:implementation-details}

\subsection{Dataset}

\begin{table}[h]
    \centering
    \begin{tabular}{l|rr}
    \toprule
        Dataset & Avg. duration & Data size \\
    \midrule
        \textbf{EgoSchema}~\cite{mangalam2023egoschema} & 180s & 500 \\
        \textbf{LVBench}~\cite{wang2024lvbench} & 4,100s & 1,549 \\
        \textbf{MMBench-Video}~\cite{fang2024mmbenchvideo} & 165s & 1,998 \\
        \textbf{VideoMME}-long~\cite{fu2024video} & 2,466s & 900 \\
    \bottomrule
    \end{tabular}
    \caption{\textbf{Dataset Statistics}. Overview of the data statistics across all four benchmarks: the minute-level EgoSchema and MMBench-Video, and the hour-level LVBench and VideoMME. For VideoMME, the statistics correspond to the \emph{long} subset.}
    \label{tab:data-statistics}
\end{table}

\begin{table}[h]
    \centering
    \begin{tabular}{p{\linewidth}}
        \toprule
        \textbf{System Prompt}  \\
        You are a helpful assistant. Please answer the following question. \\
        \midrule
        \textbf{EgoSchema}  \\
        \textcolor{task}{Taking into account all the actions performed by c, what can you deduce about the primary objective and focus within the video content?} \\
        \textcolor{task}{Option 0: C is cooking.} \\
        \textcolor{task}{Option 1: C is doing laundry.} \\
        \textcolor{task}{Option 2: C is cleaning the kitchen.} \\
        \textcolor{task}{Option 3: C is cleaning dishes.} \\
        \textcolor{task}{Option 4: C is cleaning the bathroom.} \\
        Please directly answer the option number. \\
        \midrule 
        \textbf{LVBench}  \\
        \textcolor{task}{What year appears in the opening caption of the video?} \\
        \textcolor{task}{(A) 1636} \\
        \textcolor{task}{(B) 1366} \\
        \textcolor{task}{(C) 1363} \\
        \textcolor{task}{(D) 1633} \\
        Please select the best answer from the options provided and directly provide the letter representing your choice without giving any explanation. \\
        \midrule 
        \textbf{MMBench-Video}  \\
        \textcolor{task}{What is the name of the player who scored the first goal in the video?} \\
        Please directly reply with your response. \\
        \midrule 
        \textbf{VideoMME}  \\
        \textcolor{task}{What is the video mainly about?} \\
        \textcolor{task}{A. Planes invented by the Wright Brothers.} \\
        \textcolor{task}{B. The structural difference between the planes created by Whitehead and planes created by the Wright Brothers.} \\
        \textcolor{task}{C. Who invented the first plane.} \\
        \textcolor{task}{D. How Whitehead and the Wright Brothers cooperated to invent the first motorized flight.} \\
        Please select the best answer from the options provided and directly provide the letter representing your choice without giving any explanation. \\
        \bottomrule
    \end{tabular}
    \caption{\textbf{Prompt Template Examples}. Examples of the prompt template used in our framework. Each dataset is represented by a sample instance, marked in \textcolor{task}{red}, paired with its corresponding instruction template for clarity.}
    \label{tab:prompt-example}
\end{table}

\begin{table*}[h]
    \centering
    \begin{tabular}{@{}lcrrrrrrr@{}}
        \toprule
        Method & Frames & ER & EU & KIR & TG & Rea & Sum & Avg.\\
        \midrule
        Ours & 20.0 & \textbf{43.7} & \textbf{40.7} & 37.8 & \textbf{38.0} & \textbf{46.2} & 27.3 & \textbf{41.3} \\
        - w/o Reward & 22.2 & 37.5 & 35.0 & \textbf{43.6} & 27.2 & 44.0 & \textbf{40.0} & 39.0 \\
        - w/o Tree Search & 39.9 & 34.2 & 34.7 & 40.7 & 29.4 & 41.2 & 29.1 & 36.2 \\
        \bottomrule
    \end{tabular}
    \caption{\textbf{Detailed Ablation Study Results on LVBench}. Both the tree-search exploration and reward model contribute significantly to improvements in performance and efficiency. The memory buffer size of our agent is set to \textbf{16 frames}. We report the average number of observed frames and mark the best performance in \textbf{bold}.}
    \label{exp-ablation-lvbench-detail}
\end{table*}

In this section, we elaborate the details and statistics of the benchmark datasets.
Besides the two datasets mentioned in Sec.~\ref{sec:exp-setup}, we also include the two recently released datasets, \textbf{MMBench-Video}~\cite{fang2024mmbenchvideo} and \textbf{VideoMME}~\cite{fu2024video} into our evaluation.
The statistics of the datasets are provided in Tab.~\ref{tab:data-statistics}.
The official implementation of VideoMME utilizes GPT-4 for evaluation. To ensure consistency, we also use GPT-4o as the evaluator in this work.

We acknowledge the existence of several well-established long-term video question answering benchmarks~\cite{zhang2023movqa,soldan2022mad}, such as MovieChat~\cite{song2023moviechat} and MovieQA~\cite{tapaswi2016movieqa}.
However, since the videos in these benchmarks are often sourced from publicly available movies, there is a potential risk that they may have been included in the training data of recent closed-source VLMs such as GPT-4o. To avoid data leakage and ensure fair evaluation, we chose not to use these benchmarks.

For further clarity, we provide detailed prompts of a randomly chosen example for each dataset in Tab.~\ref{tab:prompt-example}.
For each dataset, we follow the prompt template provided in their paper or official implementation.

\subsection{Baselines}

In this section, we elaborate the implementation details of the baselines.
For VideoAgent~\cite{VideoAgent} and VideoTree~\cite{wang2024videotreeadaptivetreebasedvideo}, we begin by adhering to their official implementations\footnote{\url{https://github.com/wxh1996/VideoAgent}}\footnote{\url{https://github.com/Ziyang412/VideoTree}}.
However, we notice that both methods utilize GPT-4 as the reasoning agent, which is suboptimal compared to GPT-4o, the model we employ as the exploration agent.
Therefore, to ensure a fair comparison, we re-implement both methods using GPT-4o as the reasoning agent across all benchmarks.
Furthermore, as the original implementations do not provide extracted captions for the videos, we followed their methodology by extracting captions at 0.2 FPS using the current state-of-the-art VLM, Qwen2-VL~\cite{Qwen2VL}, due to its powerful video understanding capabilities. 
\textbf{However, using GPT-4o as the captioner is computationally prohibitive, as a single 30-minute video would require over 4M tokens, making it infeasible in practice. }
For their visual encoders, we leverage EVA-CLIP-8B~\cite{sun2024eva} with the same settings as~\cite{VideoAgent,wang2024videotreeadaptivetreebasedvideo}.

For a fair comparison with VideoAgent, we experimented with varying the number of initial frames of VideoAgent to optimize its performance, accounting for changes in both the agent and caption models. The top two results on EgoSchema are presented in Tab.~\ref{exp:egoschema-main}, aligned with the optimal parameter reported in the original paper.
For LVBench and Video-MME, we set the initial frame count to 15 to better accommodate the processing requirements of longer videos. For the clustering component of VideoTree, we provide the hyper-parameter settings as follows: \text{max\_breadth} = 32, \text{max\_depth} = 3, \text{branch\_width} = 4, and \text{rele\_num\_thresh} = 4.





\section{Experiments}
\label{app:experiments}

\subsection{Experimental Results}

\begin{table}[t]
    \centering
    \begin{tabular}{@{}lrrr@{}}
        \toprule
        Method & VideoAgent$^\dagger$ & VideoTree$^\dagger$ & Ours \\
        \midrule
        Avg. Frames & 24.6 & 98.0 & \textbf{18.1} \\
        \midrule
        Knowledge & 52.2 & \textbf{60.7} & 56.9 \\
        Film \& Television & 42.5 & 52.5 & \textbf{55.0} \\
        Sports Competition & 42.7 & 48.6 & \textbf{59.3} \\
        Artistic Performance & 47.5 & 51.6 & \textbf{65.8} \\
        Life Record & 44.7 & 49.5 & \textbf{51.9} \\
        Multilingual & 36.6 & 40.0 & \textbf{46.7} \\
        \midrule
        Overall &46.4  & 53.1 & \textbf{56.3} \\
        \bottomrule
    \end{tabular}
    \caption{\textbf{Experimental Results on VideoMME Long Split}. Methods marked with $^\dagger$ indicate implementations where \textbf{GPT-4o} serves as the main component. The memory buffer size of our agent is set to \textbf{16 frames}. We report the average number of observed frames and mark the best performance in \textbf{bold}.}
    \label{exp:videomme-main}
\end{table}

As discussed in Sec.~\ref{sec:exp-setup}, we present the experimental results on VideoMME and MMBench-Video in Tab.~\ref{exp:videomme-main} and Tab.~\ref{exp:mmbench-main}, respectively.
For VideoMME, we evaluate our framework on the long split, with videos averaging over 2,000 seconds in duration.
Similarly, to ensure a fair comparison, we re-implement the baselines using GPT-4o as the base VLM.
As shown in Tab.~\ref{exp:videomme-main}, our method achieves a significant improvement, outperforming VideoTree by 4.9\% with less than 20\% of observed frames, and outperforming VideoAgent by 12.1\% with about 75\% of observed frames. Similar trends are observed in the results for MMBench-Video in Tab.~\ref{exp:mmbench-main}.
We observe that both VideoAgent and VideoTree perform poorly on this benchmark. We conjecture that this is due to the long split’s emphasis on detailed visual clues, while captioning-based methods inherently struggle with tasks requiring detailed visual comprehension.



\begin{table}[t]
    \centering
    \scalebox{0.9}{
    \begin{tabular}{@{}lrrrr@{}}
        \toprule
        Method & GPT-4o & VideoAgent$^\dagger$ & VideoTree$^\dagger$ & Ours \\
        \midrule
        Avg. Frames & 8 & 7.8 & 27.1 & 7.4 \\
        Score & 1.62 & 1.05 & 1.38 & \textbf{1.68} \\
        \bottomrule
    \end{tabular}
    }
    \caption{\textbf{Experimental Results on MMBench-Video}. Methods marked with $^\dagger$ indicate implementations where \textbf{GPT-4o} serves as the main component. The memory buffer size of our agent is set to \textbf{8 frames}. We report the average number of observed frames and mark the best performance in \textbf{bold}.}
    \label{exp:mmbench-main}
\end{table}

\subsection{Ablation Study}

In this section, we present the detailed results of the ablation study discussed in Sec.~\ref{sec:exp-ablation}.
The results across different domains of LVBench are summarized in Tab.~\ref{exp-ablation-lvbench-detail}.
Overall, the results demostrate that both the reward model and the tree-search exploration mechanism play significant roles in enhancing performance. Interestingly, we observe that the agent without the reward model or tree-search exploration performs better in the Key Information Retrieval (KIR) and Summarization (Sum) domains. This result is unsurprising, as the absence of tree-search exploration prompts the agent to explore nearly twice as many frames. Similarly, without the reward model, the agent lacks focused guidance and instead explores more broadly across the video.

Conversely, these components contribute significantly to performance improvements in tasks like Event Recognition (ER) and Temporal Grounding (TG), where precise and focused exploration is essential. Overall, the reward model and tree-search exploration, each addressing distinct aspects of the exploration process, work together to drive large performance gains across diverse tasks, aligning with our findings in Sec.~\ref{sec:exp-ablation}.

\clearpage

\onecolumn

\section{Case Study}
\label{app:case-study}

As mentioned in Sec.~\ref{sec:case-study}, in this section, we investigate the common failure cases of our framework, aiming to provide data points and insights for the future research.

\subsection{Failure Mode: Inability to Detect Subtle Visual Details}

\begin{figure*}[h]
\centering
\centerline{\includegraphics[width=0.8\linewidth]{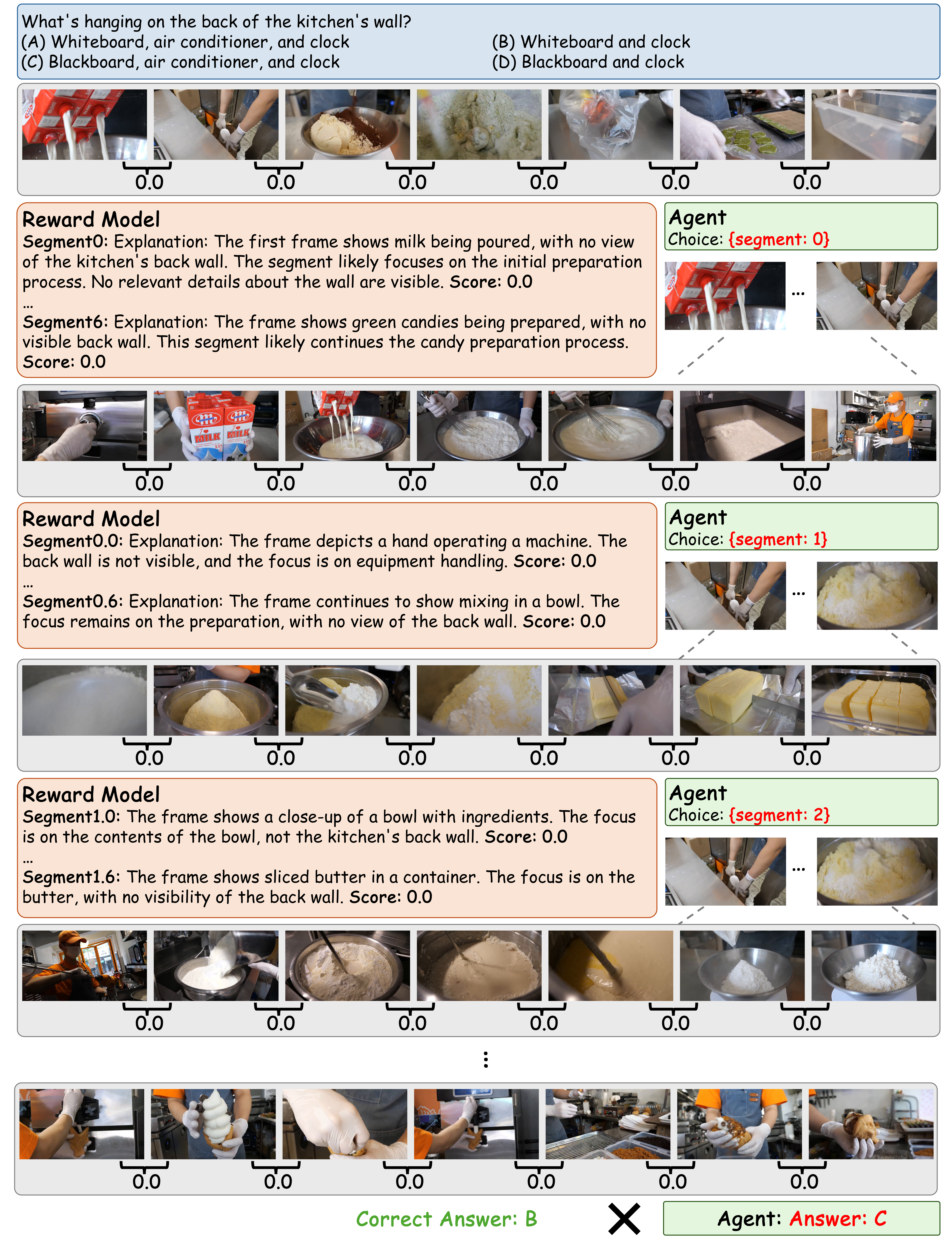}}
\caption{\textbf{Failure Exploration Trajectory Example}. The agent struggles to locate crucial information for questions requiring visual details, even after prolonged exploration, ultimately resulting in a random generated answer.}
\label{fig:bad-case-1}
\end{figure*}

The most common error occurs when the agent overlooks subtle clues and fails to extract crucial information, even after extensive exploration. A representative example is illustrated in Fig.~\ref{fig:bad-case-1}. In an egocentric cooking video, the agent is tasked with identifying decorations on the back wall of the kitchen, details that appear only briefly during rapid camera movements, making the task challenging even for humans. After multiple rounds of exploring irrelevant segments, the agent fails to locate the key frames and ends up with terminating exploration, and provides a guessed answer. In such cases, the number of frames observed is typically three times the average. We attribute these errors to the inherently high difficulty of questions including subtle visual details, an extremely tricky task for humans as well.

\subsection{Failure Mode: Guidance Errors from the Reward Model}

\begin{figure*}[h]
\centering
\centerline{\includegraphics[width=0.9\linewidth]{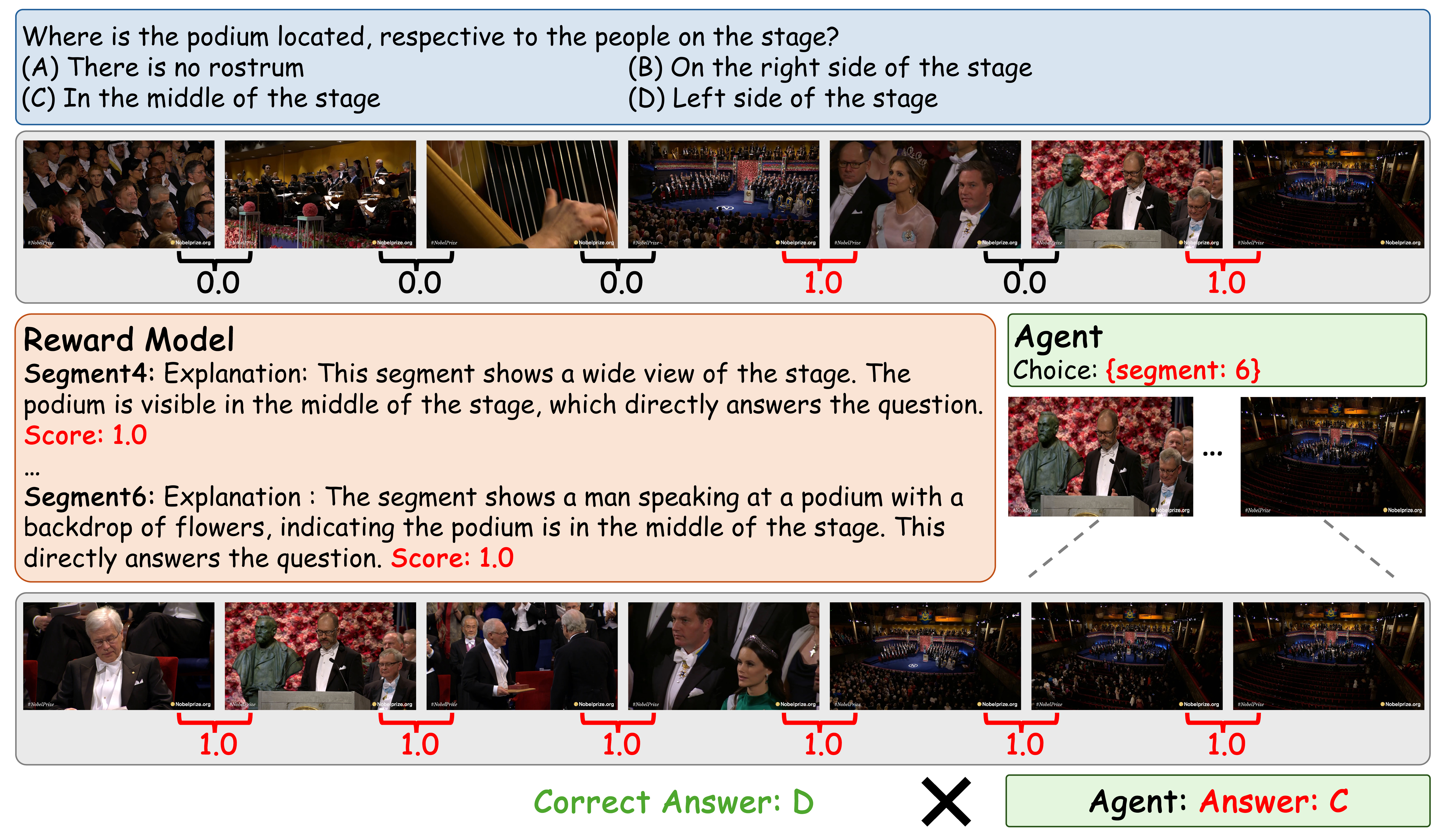}}
\caption{\textbf{Failure Exploration Trajectory Example}. The agent is misled by inaccurate relevance scores from the reward model, causing it to focus on incorrect segments and provide an inaccurate answer.}
\label{fig:bad-case-2}
\end{figure*}

Another common failure mode occurs when the agent is misled by the reward model, as demonstrated in Fig.~\ref{fig:bad-case-2}. For instance, when tasked with identifying the spatial relationship between the podium and the stage, the reward model incorrectly assigns 100\% relevance scores to the 4th and 6th segments, while the correct information lies in the 5th segment. Consequently, the agent focuses on the wrong segments and provides an incorrect answer. These errors stem from the inherent limitations of the reward model, which fails to provide robust and accurate guidance for the exploration agent. This limitation is also discussed in Sec.~\ref{sec:reward-quality}.

\subsection{Failure Mode: Limited Multi-modal Reasoning Abilities}
\begin{figure*}[h]
\centering
\centerline{\includegraphics[width=0.9\linewidth]{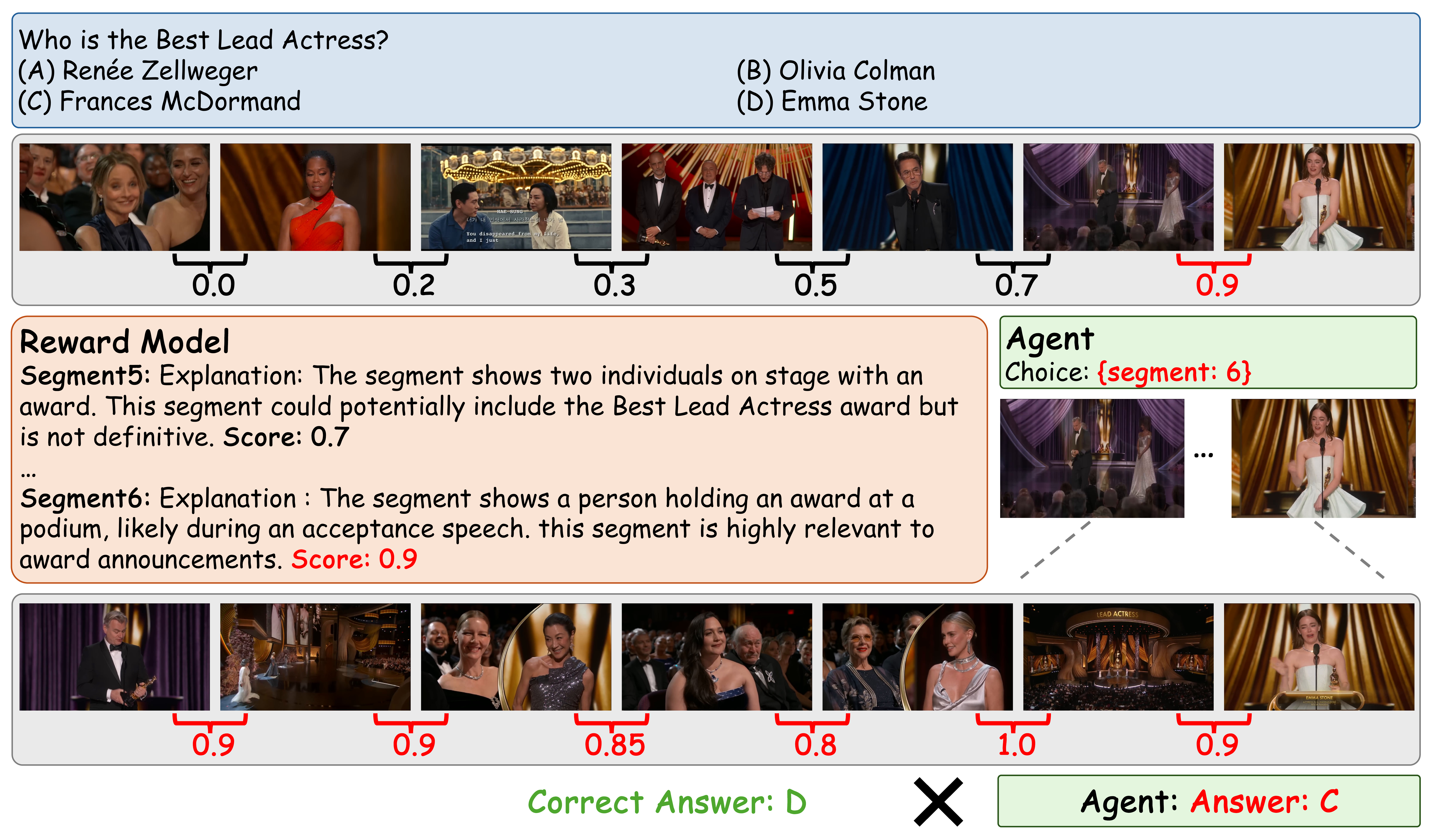}}
\caption{\textbf{Failure Exploration Trajectory Example}. The agent successfully identifies the ground truth visual information but fails to deliver the correct response.}
\label{fig:bad-case-3}
\end{figure*}

Another typical failure arises when the agent successfully identifies the crucial segments but fails to produce the correct answer due to limited spatial reasoning capabilities or inadequate alignment of multi-modal knowledge. For example, as shown in Fig.~\ref{fig:bad-case-3},  the agent correctly identifies the frame indicating the winner of the Best Lead Actress, with the name clearly annotated in the subtitles. However, even when provided with ground truth visual information, the agent is unable to generate the correct response. We attribute these errors to the inherent limitations of the exploration agent, which suggests the potential for further improving our framework by integrating more advanced foundation models.

\subsection{Discussion}
The analysis of failure modes reveals the large potential of our framework. While detecting subtle visual details remains challenging even for human, the error stemming from guidance failures by the reward model and reasoning failures by the exploration agent could be mitigated by incorporating more robust multi-modal foundation models. As these foundational models evolve, our framework can seamlessly integrate these improvements, indicating its strong potential for tackling more challenging tasks in long-video understanding.